# Small Celestial Body Exploration with CubeSat Swarms

Emmanuel Blazquez[1,a *], Dario Izzo[1,b], Francesco Biscani[2], Roger Walker[3] and Franco Perez-Lissi[3]

[1]European Space and Research Technology Centre, Advanced Concepts and Studies Office, European Space Agency, Kepleerlan 1, 2201AZ Noordwijk (The Netherlands)

[2] European Space Operations Centre, , European Space Agency, Robert-Bosch-Straße 5, 64293 Darmstadt (Germany)

[3] European Space and Research Technology Centre, Cubesat Systems Unit, European Space Agency, Kepleerlan 1, 2201AZ Noordwijk (The Netherlands)

[a]emmanuel.blazquez@esa.int, [b]dario.izzo@esa.int

* corresponding author

**Abstract.** This work presents a large-scale simulation study investigating the deployment and operation of distributed swarms of CubeSats for interplanetary missions to small celestial bodies. Utilizing Taylor numerical integration and advanced collision detection techniques, we explore the potential of large CubeSat swarms in capturing gravity signals and reconstructing the internal mass distribution of a small celestial body while minimizing risks and Delta V budget. Our results offer insight into the applicability of this approach for future deep space exploration missions.

**Introduction**

In the last decade CubeSats have emerged as an innovative and cost-effective platform for testing new satellite technologies, with applications ranging from Earth observation to deep space missions. For instance, the HERA interplanetary mission that will explore the Didymos binary system in 2025 will embark two CubeSats to perform a detailed exploration of the system. [1] In this context, distributed swarms of CubeSats are a promising strategy for future exploration missions to small celestial bodies, enabling increased scientific return while minimizing risks associated with operating in an unknown environment. [2]

In this work, we present a large-scale simulation study of the deployment and operation of many CubeSats around small celestial bodies, with the aim of assessing the applicability of large CubeSats swarms for distributed operations during interplanetary missions to asteroids and comets. Our approach leverages the *cascade* and *heyoka* C++/Python libraries to propagate the evolution of the swarm, reliably detecting close encounters and collisions using high-order Taylor numerical integration and collision detection. [3] We assume that the swarm is deployed sequentially by a "mother" spacecraft operating in proximity of the celestial body of interest, and that each CubeSat has the capability of performing 6-DoF orbital manoeuvres with limited ΔV budget. We use mascon mass models to have a representation of the mass distribution of the bodies around which the swarm is orbiting and consider CubeSat deployment uncertainty in position and velocity as well as a fully automated trajectory control strategy aimed at keeping the overall Delta V budget under control while maximizing scientific return and controlling the risk of conjunctions. Our simulations offer insights into the trade-offs involved in deploying and maintaining large CubeSat swarms for distributed operations in deep space.

**Background and Models**
In this study, we consider a prospective advanced mission concept inspired by the European Space Agency's HERA mission. [1] We assume that a mothership satellite is inserted into orbit around a small celestial body of irregular shape and will consequently deploy a swarm of CubeSats from a common orbit. The swarm's objective is to capture gravity signal around the body to assess its mass distribution at a later stage following the procedure described in the *geodesyNETs* project. [4] The swarm CubeSats will have to orbit around the body without colliding with each other, staying within a spherical region defined by a minimum safety radius to the body's center of mass and a maximum radius of operations. These operations will have to be performed with minimum ΔV budget requirements.

Small celestial bodies of interest for this study will have their internal mass distribution represented by a heterogeneous mascon model. In other words, they will be modelled as a set of point masses located in specific positions and with some given mass. The mascon models used for this study were developed in the *geodesyNETs* project. [4] The dynamics of the swarm around the spacecraft are simulated via cascade simulations making use of the *heyoka* high-performance Taylor integrator. The equations of motion are represented by a gravitational *N*-body problem where *N* is the number of mascons, plus the solar radiation pressure integrated in an inertial reference frame and consider a rotating celestial body. Terminal events are added to the equations of motion to enforce a maneuver when a spacecraft enters the safety sphere around the body or exits the operations sphere.

We assume that each CubeSat of the swarm has the capability of performing maneuvers with full degrees of freedom whenever it gets close to another element of the swarm, to the central body or flies too far away. A 2-phase collision detection algorithm is used to detect when two elements of the swarm get too close to each-other. It exploits the availability, at each step, of Taylor polynomial expansions of the system dynamics that are the product of the integration scheme offered by *heyoka* and therefore come at no additional cost. *Cascade* makes use of these expansions to compute Axis Aligned Bounding Boxes (AABB) encapsulating the positions of each orbiting object within a chosen collisional timestep. The algorithm operates in two phases: a broad phase with a Bounding Volume Hierarchy (BVH) over the 4-dimensional AABB defined in cartesian coordinates, followed by a narrow phase more computationally demanding making use of a polynomial root finding algorithm. This approach is well-suited to parallelization and particularly performant in non-collision rich environments such as the ones considered in this study.

**Simulations and Results**
We apply our framework to simulate swarm missions to Itokawa, Bennu and Ryugu. [5,6,7] For each body four swarm configurations are considered, with 5, 10, 25 and 50 spacecraft respectively. Itokawa is modelled using 24800 mascons, Bennu with 17400 mascons and Ryugu with 26800 mascons. The mothership is initially located on a circular orbit around the body of interest with a semi-major axis of 1.5km, an inclination of 90° and a mean anomaly of 90°. The solar radiation pressure acting on the swarm spacecraft is computed considering a wet area of 1 m$^2$, with the sun direction being selected arbitrarily in the ecliptic plane. The spacecraft must remain in an ellipsoidal region around the asteroid, with a no-entry safety ellipsoid whose dimensions are those of the celestial body with a factor 1.3 and a no-exit sphere of radius 3 km for Itokawa and 2 km for Bennu and Ryugu. The relative release velocities of the swarm spacecrafts with respect to the mothership are randomly selected in [1.5, 3.5] cm/s. 100 collisional timesteps are processed in parallel to account for CubeSat collisions, where one timestep is defined as $1/60^{th}$ of the orbital period of the mothership spacecraft and is also representative of the release frequency of the CubeSats in the swarm. The collisional radius, the minimum relative distance required between two CubeSats to avoid collisions, is set to 5 m. The dynamics and operations of the swarm are simulated in each case for a total duration of 4 days.

**Table 1:** *Simulation results for swarms orbiting around Itokawa/Bennu/Ryugu,*

|  | Itokawa | | | | Bennu | | | | Ryugu | | | |
|---|---|---|---|---|---|---|---|---|---|---|---|---|
| **Swarm size** | 5 | 10 | 25 | 50 | 5 | 10 | 25 | 50 | 5 | 20 | 25 | 50 |
| **ΔV$_{min}$ [m/s]** | 0.1 | 0.1 | 0.1 | 0.1 | 0 | 0 | 0 | 0.03 | 0 | 0 | 0 | 0 |
| **ΔV$_{max}$ [m/s]** | 1.4 | 2.6 | 3 | 3.7 | 0,66 | 2.9 | 2.6 | 4.5 | 0.26 | 0.26 | 1.04 | 1.63 |
| **ΔV$_{mean}$ [m/s]** | 0.82 | 0.85 | 0.97 | 1.4 | 0.39 | 0.68 | 0.7 | 0.92 | 0.08 | 0.05 | 0.12 | 0.34 |
| **Collision events** | 2 | 2 | 5 | 12 | 2 | 9 | 51 | 89 | 4 | 4 | 21 | 77 |
| **Safety events** | 7 | 9 | 39 | 191 | 34 | 15 | 40 | 78 | 4 | 4 | 12 | 36 |
| **Re-entry events** | 39 | 82 | 192 | 393 | 46 | 114 | 259 | 481 | 4 | 9 | 26 | 142 |

Table 1 showcases the Results obtained from the simulations. The number of events requiring maneuvers is in general dominated by re-entry requirements when the satellites are in danger of exiting their operational region around the asteroid. As expected, the size of the swarm has a direct impact on the ΔV budget for the mission. The increase observed in the maximum ΔV required for swarms with a higher number of spacecrafts is partially due to the increased number of collisional events recorded, especially for Bennu and Ryugu. However, re-entry events account for most of the budget and increase proportionally with the number of satellites in the swarm. This is because, on average, every CubeSat of the swarm will need to be actuated several times to stay within the operational region for the mission. Safety events are more predominant for Itokawa due to the elongated shape of the body that makes trajectory correction maneuvers more likely to cause future safety events.

Of particular interest is the fact that the trade-off for the number of satellites in the swarm, therefore gravity signal recovery, versus the ΔV budget of the mission is dependent on the shape and mass distribution of the body. Itokawa sees an 18% increase in mean ΔV going from 5 to 25 satellites in the swarm, for a drastic increase in recovered gravity signal and therefore internal shape reconstruction. But the increase from 25 to 50 satellites leads to a 44% increase in mean ΔV: the number of collision events increases enough to create a cascade of safety events that increase the overall budget significantly. The tradeoff for Bennu and Ryugu appears at different sizes for the swarm but exhibits a similar behavior: an increased number of collisional events leads to an increased number of safety and re-entry events and therefore hinders the mission budget. This could signify that the number of collision events dominates the trade-off analysis and could be a suitable preliminary indicator to size the swarm when gravity signal reconstruction is sufficient. Regularity in the shape of the body seems to be correlated with a higher number of collisional events but overall lower maintenance budgets, as the cascade phenomenon is not as pronounced. This analysis will be refined in future work with a more robust optimization setup, the purpose of this work being principally to propose and make available a framework for the efficient parallel simulation of CubeSat swarm operations around small celestial bodies.

Fig. 2 presents a top-view of a 50-spacecraft swarm simulation orbiting Itokawa as well as the asteroid shape reconstruction from the gravity signal gathered from the swarm after 4 days of operations and using the procedure described in the *geodesyNETs* project, along with the ΔV budget distribution for the swarm. [4]

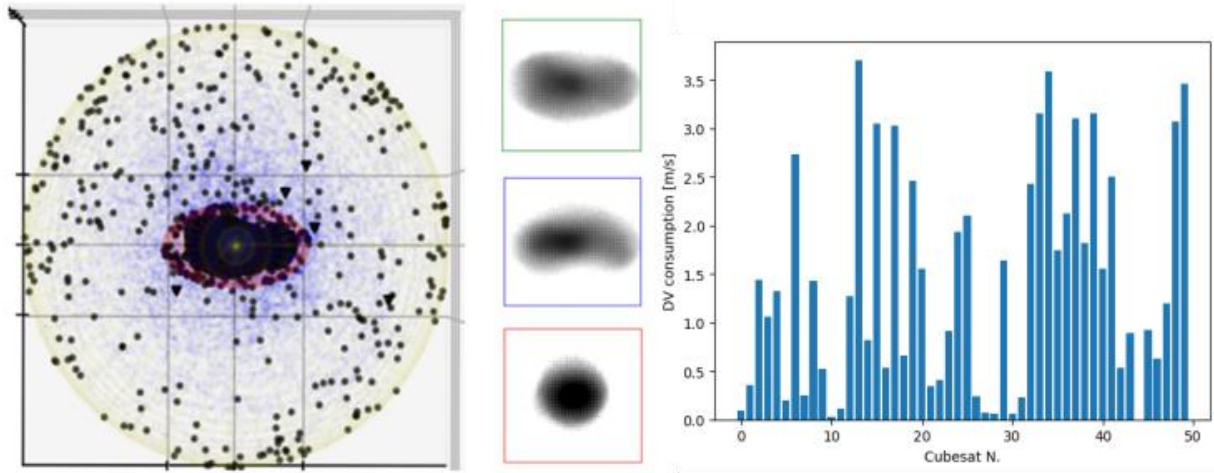

*Figure 2*: *Itokawa study case, (left) simulation of a collision-free swarm of 50 elements. The position of the swarm elements during the gravity field measurements are shown in blue. Impulsive maneuvers to keep the swarm in safe proximity are shown in black (collision avoidance, triangles). (middle) reconstruction of the asteroid shape after 4 days of operation. (right) ΔV budget distribution across the CubeSats in the swarm.*

**Conclusion**

In this work we have applied our framework to simulate CubeSat swarm missions to Itokawa, Bennu and Ryugu. We show how swarms of different sizes, ranging from a few CubeSats to several dozen, can efficiently capture gravity signal and reconstruct spherical harmonic expansions while minimizing the risk of conjunction and counteracting the effects of environmental disturbances. Our results showcase the potential of large CubeSat swarms for future exploration missions to small celestial bodies and propose an Open-source framework for preliminary mission analysis studies in such scenario. The code used for this study will be made publicly available at https://gitlab.com/EuropeanSpaceAgency/collision-free-swarm, and the *cascade* library used for this work is an open source python module available at https://github.com/esa/cascade.